\documentclass[conference]{IEEEtran}
\usepackage{graphicx}
\ifCLASSINFOpdf
\else
\fi
\begin{document}
%
\title{Transfer Learning for Credit Card Fraud Detection: A Journey from Research to Production}

\author{\IEEEauthorblockN{Wissam Siblini, Guillaume Coter, Rémy Fabry \\
Liyun He-Guelton \& Frédéric Oblé}
\IEEEauthorblockA{\textit{Worldline}\\
\textit{Brussels, Belgium \& Lyon, France}\\
\textit{Email: wissam.siblini@worldline.com}}
\and
\IEEEauthorblockN{Bertrand Lebichot, Yann-Aël Le Borgne \\
\& Gianluca Bontempi}
\IEEEauthorblockA{\textit{Universite Libre de Bruxelles}\\
\textit{Brussels, Belgium}\\
\textit{Email: bertrand.lebichot@ulb.ac.be}}}


%


\maketitle

\begin{abstract}
The dark face of digital commerce generalization is the increase of fraud attempts. To prevent any type of attacks, state-of-the-art fraud detection systems are now embedding Machine Learning (ML) modules. 
The conception of such modules is only communicated at the level of research and papers mostly focus on results for isolated benchmark datasets and metrics. But research is only a part of the journey, preceded by the right formulation of the business problem and collection of data, and followed by a practical integration. In this paper, we give a wider vision of the process, on a case study of transfer learning for fraud detection, from business to research, and back to business.
\end{abstract}


%
\IEEEpeerreviewmaketitle

\section{Introduction}

  Financial institutions that process payment have to fight a tremendous amount of fraudulent activities. In 2018, the level of card fraud losses amounted to 1.8 billion euros in the Single European Payment Area \cite{ecb2018}. On a global scale, it is expected to reach more than 34 billion by 2022. Payment processors also face indirect repercussions as fraud jeopardizes the trust of customers, namely banks and cardholders. It is therefore obvious for them to resort to Fraud Detection Systems (FDS). Most FDS has a rule-based basis (e.g. "raise an alert if the last transaction from cardholder occurred less than 5 seconds before") \cite{kou2004survey}, designed by experts, which can be specific (low number of false positives) but not very sensitive nor flexible. The cat-and-mouse game played by fraudsters makes them quickly obsolete, and their update requires a lot of manual work. Adding an incremental ML-based system can provide complementarity (from experience, combining expert rules and ML models provides from 20\% to 100\% better results than each approach alone) and adaptability to make an overall powerful FDS \cite{dal2014learned,lebichot2020incremental}. ML also opens the door to underlying techniques like Transfer Learning (TL) \cite{lebichot2021transfer}, which is essential when starting working on a new payment dataset/domain (e.g. new bank) with very few annotations. TL allows adapting an existing efficient model to use it on the new domain. This provides a competitive business advantage in terms of customer prospects by promising a less perilous start. 
  
  In this paper, we give an overview of a 3-year project between Worldline and ULB, involving a dozen researchers. We focus on a TL study for FDS, carried out over two years for research and one year for industrialization, with a 6-month overlap. Rather than detailing our research findings (available in \cite{lebichot2021transfer}), we describe the back-and-forth journey between business and production: from the formulation of practical business constraints to the research study definition and conclusions, and finally the transformation into concrete assets for deployment, including our return on experience about reproducibility issues and how to anticipate them. 

\section{The Business constraints and challenges}

ML research for FDS starts with open questions (e.g. how to detect new fraud patterns quickly?). Then, one selects an appropriate dataset, proposes an algorithm to optimize adequate performance metrics. But having to implement the results in production imposes a lot of additional constraints: (1) \textbf{Data collection} must be done with caution (e.g. proper train-test temporal split with verification gap \cite{dal2014learned}, denoised annotations, etc.). Such data-centric considerations are now acknowledged by experts to be key in the overall ML process. \cite{alazizi2019anomaly} gives a non-exhaustive list of those related to the fraud problem. (2) \textbf{Metrics}: Apart from detection rate, metrics like memory/time consumption are primordial for efficient real-time fraud detection \cite{tran2018real}. Contracts can also impose a minimum level of performance (e.g a precision above 1/3). Finally, interpretable indicators are required for continuous improvement \cite{siblini2020master}. (3) \textbf{Compliance and interoperability}: the ML pipeline should comply with regulations (e.g. limits in data storage/usage), be integrable in the payment process, and interact properly with the production platform and services. (4) \textbf{Relatively to transfer learning}, specific constraints can arise like having to homogenize the variables encoding and the processing pipeline between source and target domains.

\section{The research problem and its conclusions}

In Transfer Learning, we consider domains $D$ (described by variables $X$ and distributions $P(X)$) and tasks $T$ (e.g. predict fraud $y$ by modeling $P(y|X)$ with a function $f$). Given a source ($D_s$,$T_s$) with a rich dataset, TL adapts the associated solution $f_s$ to compute a solution $f_t$ for a target ($D_t$,$T_t$) where the dataset is poor in annotations. In our project, there are two use-cases of TL: reusing an FDS trained on online payment (resp. payment in country A), for payment on physical terminals (resp. in country B). These cases fall into a category called "Homogeneous domain adaptation": same tasks $T_s=T_t$ and close sets of variables, but different distributions since customers (resp. fraudsters) do not have the same habits (resp. techniques). In our experiments, the source (resp. target) set was made of around 140 (resp. 60) million transactions over 183 operating days, with a fraud ratio in the order of 0.1\%. A transaction is described by around 20 features (e.g. amount, country, payment method, merchant, cardholder). Our work consisted in adapting, implementing, and evaluating a large number of approaches from the literature: standard base models (NB, RF, NN, Xgboost) with several transfer strategies (model trained on the source or target or both, transfer with domain-invariant representations, etc.). For the sake of conciseness, we refer the reader to \cite{lebichot2019deep,lebichot2021transfer} for more details and results. We concluded that models trained on source perform better when the target domain is poorly labeled. When target labels get less scarce, semi-supervised domain adaptation methods take over the reins. Overall, a combination of a \textit{model trained only on source} and a \textit{model trained on both target and source} performs well in most cases. This combination is under integration into production, as it allows a global improvement of detection by around 15\% compared to a zero-shot transfer baseline. With thousands of fraud attempts every day, this allows significant financial savings.






\section{The Integration Methodology: make it Happen}

\begin{figure}[h]
\centering
\includegraphics[width=0.75\columnwidth]{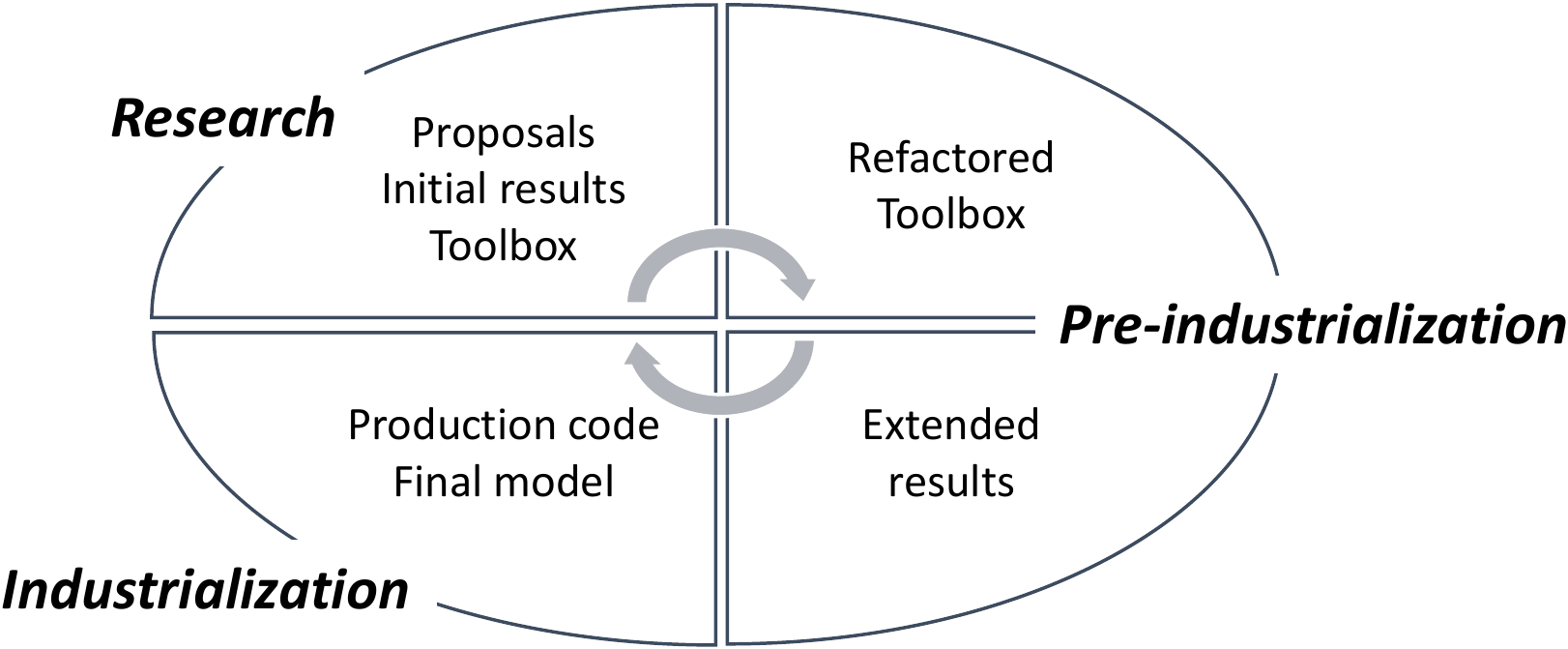}
\caption{A cyclic industrialization process.}
\label{fig:IndustrializationProcess}
\end{figure}

Figure~\ref{fig:IndustrializationProcess} shows the steps to industrialize findings. \textbf{Research} is followed by \textbf{Pre-industrialization}, during which (1) the work is adapted to be integrated into production-compatible tools (libraries, versioning tools) and (2) results are confirmed on larger datasets. Then comes \textbf{Industrialization}, where methods are adapted and deployed to the production environment.

As the steps are usually carried one after another (with delays of several months) and have their own requirements, reproducibility issues arise. Some are related to \textbf{data}. Usually, a rather small subset of data is selected for the research part. Meanwhile, new data is collected so when the study is over, one has to check if results still hold. Sometimes, it is necessary to change the data format and integrate new features. Other issues are related to the \textbf{code}. Research and pre-industrialization use different environments (also different from the actual production environment) due (i) to their different objectives (research can require a flexible environment like \textit{Jupyter}) and (ii) due to collaboration between different entities (e.g. academics and industrials). As a consequence, library versions require adaptation either in the delivered code or in the platform welcoming it. Additionally, a research implementation contain code snippets that are already implemented in production. In this case, the code needs to be refactored accordingly. Lastly, versioning tools (Git, DVC, wandb, ...) for code, data, and results are essential in later steps. To properly apply them, the code needs adaptations as well. 

To tackle reproducibility challenges, our main return on experience is that all steps of the project should not be seen as linear and successive. It is important (1) to anticipate: providing pieces of the production environment as a basis for the research implementation allows an easier later integration (savings counted in months) and even a cleaner formulation of the problem; and (2) to do several quick rounds with all the steps instead of a single long one: integrating code soon allows to capture hidden issues and redefining formulation actively to reorientate the study. However, it remains important to maintain flexibility because when too restricted by business constraints, findings can remain limited to incremental innovations with moderate added value \cite{assink2006inhibitors}.

\section{Conclusion}

Research and development are important for competitiveness in the industry. It demonstrates expertise and creates value. It is also at the core of the ML revolution. But we usually tend to only communicate on the research part of the process. Yet the complete path from emergence to deployment hides many other challenges. A good practice is to anticipate them to go further faster. Promising formalized directions start to emerge for instance with MLOps, the set of recipes/tools for ML deployment. This however does not mean that businesses should constrain research. It allows efficient integration of course but can prevent disruptive innovations \cite{assink2006inhibitors}.




\bibliographystyle{IEEEtran}
\bibliography{biblio.bib}
%

\end{document}